\documentclass[conference]{IEEEtran}

\usepackage{amsmath}
\usepackage{amssymb}
\usepackage{amsthm}
\newtheorem{definition}{Definition}

\usepackage{multirow}
\usepackage{booktabs}
\usepackage[pdftex]{graphicx}

\begin{document}

\title{Learning Pairwise Disjoint Simple Languages \\ from Positive Examples}

\author{
\IEEEauthorblockN{Alexis Linard\IEEEauthorrefmark{1}, Rick Smetsers\IEEEauthorrefmark{1}, Frits Vaandrager\IEEEauthorrefmark{1}, Umar Waqas\IEEEauthorrefmark{2}, Joost van Pinxten\IEEEauthorrefmark{2} and Sicco Verwer\IEEEauthorrefmark{3}} 
\IEEEauthorblockA{\IEEEauthorrefmark{1} Institute for Computing and Information Science, Radboud University, Nijmegen, The Netherlands\\ Email: \{a.linard, r.smetsers, f.vaandrager\}@cs.ru.nl}
\IEEEauthorblockA{\IEEEauthorrefmark{2} Department of Electrical Engineering, Eindhoven University of Technology, The Netherlands \\ Email: \{u.waqas, j.h.h.v.pinxten\}@tue.nl}
\IEEEauthorblockA{\IEEEauthorrefmark{3} Department of Intelligent Systems, Delft University of Technology,\ The Netherlands\\ Email: s.e.verwer@tudelft.nl}
}

\maketitle

\begin{abstract}
A classical problem in grammatical inference is to identify a deterministic finite automaton (DFA) from a set of positive and negative examples.
In this paper, we address the related -- yet seemingly novel -- problem of identifying a \emph{set} of DFAs from examples that belong to \emph{different} unknown simple regular languages.
We propose two methods based on compression for clustering the observed positive examples.
We apply our methods to a set of print jobs submitted to large industrial printers.
\end{abstract}

\section{Introduction}

A classical problem in grammatical inference is to find, i.e.\ \emph{learn}, a regular language from a set of examples of that language.
When this set is divided into positive examples (belonging to the language) and negative examples (not belonging to the language), the problem is typically solved by searching for the smallest deterministic finite automaton (DFA) that accepts the positive examples, and rejects the negative ones.
Provided with sufficiently many examples, there exist algorithms that will correctly learn the unknown language from these examples \cite{Gold1967}. 
This is not necessarily the case, however, if only positive examples are available.

We consider a setting where one can observe positive examples from multiple different regular languages, but it is not clear to which of these languages the examples belong.
Moreover, it is known that each example belongs to exactly one of these languages, i.e.\ that the languages are disjoint.
As a result, the positive examples for one language are negative examples for the other languages.
The problem at hand is to cluster the examples by assigning them to a language.

In this paper, we present two clustering approaches based on an essential property of regular languages described by the pumping lemma.
The pumping lemma is often used to prove that a certain language is \emph{not} regular.
It states that any sufficiently long example from a regular language has a middle section that may be repeated (i.e. \emph{pumped}) to produce a new example that is also in that language \cite{rabin1959finite}.
Efficient algorithms can be used for finding such possible decompositions \cite{gusfield2004linear}.
We use this set of possible decompositions for clustering our examples.
The intuition behind this approach is that examples from the same language are more likely to have observable similarities than examples from different languages.

Our problem is motivated by a case study of print jobs submitted to large industrial printers.
Strings of symbols represent these print jobs, where each symbol denotes a different media type, such as a book cover or a newspaper page.
Together, this set of print jobs make for a fairly complicated `language'.
Nevertheless, we observed that each print job can be classified as belonging to one of a fixed set of categories, such as `book' or `newspaper'.
Two print jobs that belong to the same category are typically very similar, to the extent that they only differ in the number of repetitions of a certain media type.
Therefore, the languages for the different categories are fairly simple.
Our goal is to uncover these simple languages.

\section{Preliminaries}
\label{sec:preliminaries}

\paragraph{Strings}
Let $\Sigma$ denote a finite alphabet of \emph{symbols}. 
A \emph{string} is a finite sequence of symbols.
The empty string is denoted $\epsilon$.
We denote with $\Sigma^{\ast}$ the set of all strings over $\Sigma$, and with $\Sigma^{+}$ all nonempty strings over $\Sigma$ (i.e.\ $\Sigma^{\ast} = \Sigma^{+} \cup \{ \epsilon \}$).
Similarly, we denote with $\Sigma^i$ the set of all strings over $\Sigma$ of length $i$.
Let $x$ be a string over $\Sigma$, then $|x|$ denotes the length of $x$.
We denote with $x^i$ the string $x$ repeated $i$ times, e.g.\ $x^2 = xx$.
We use $x^{\geq i}$ to denote the language $\{x^j : j \geq i\}$.
Moreover, we use $x^+$ to denote the language $\{x^i : i \in \mathbb{N}^+\}$, and $x^*$ to denote the language $x^+ \cup \{\epsilon\}$.
A string $u$ is a (proper) \emph{prefix} of $x$ if there exists a (nonempty) string $v$ such that $x = uv$.
Similarly, $u$ is a (proper) \emph{suffix} of $x$ if there exists a (nonempty) string $v$ such that $x = vu$. 
We say $u$ is a (proper) \emph{infix} of $x$ if there exist (nonempty) strings $v, w$ such that $x = vuw$.
In the remainder of this paper, we use $a$, $b$ and $c$ to refer to symbols in $\Sigma$, and we use $r \ldots z$ to refer to strings over $\Sigma$.

\paragraph{Languages}
A \emph{language} is a set of strings.
A regular language can be described by a \emph{deterministic finite automaton} (DFA), which is a tuple $(\Sigma, Q, q_0, \delta, F)$, where $Q$ is a set of \emph{states}, $q_0 \in Q$ is the \emph{start state}, $\delta: Q \times \Sigma \to Q$ is a \emph{transition function} from states and symbols to states, and $F \subseteq Q$ is a set of \emph{final states}.
The set of tails described by a state $q$ is the set $\{ x \in \Sigma^* : \delta(q, x) \in F\}$.
Hence, the set of tails of the start state of a DFA describes its language.
Let $q$ be a state, $a$ be a symbol and $u$ be a string, then we extend $\delta$ to $\delta^{\ast}: Q \times \Sigma^{\ast} \to Q$ by $\delta^{\ast}(q, \epsilon) = q$ and $\delta^{\ast}(q, au) = \delta^{\ast}(\delta(q, a), u)$.
We slightly abuse notation and simply use $\delta$ to refer to $\delta^{\ast}$ in the remainder of this paper.

\paragraph{The pumping lemma}
The pumping lemma \cite{rabin1959finite} states that all sufficiently long strings of a regular language contain an infix that may be \emph{pumped}, i.e.\ repeated an arbitrary number of times to construct a new string that is in the language.  

Let $\mathcal{L}$ be a regular language, then there exists an integer $p \geq 1$ depending only on $\mathcal{L}$ such that every string $x \in \mathcal{L}$ of length at least $p$ can be written as $x = uvw$ such that $|v| \geq 1$, $|uv| \leq p$ and $\forall i \geq 0,\ uv^iw \in \mathcal{L}$.
The intuition is that $v$ will always lead back to the same state, i.e.\ $\delta(q_0, u) = \delta(\delta(q_0, u), v) = q$. Therefore, a string $uv^iw$ is in the language iff $w$ is in the set of tails of $q$.

We say that two strings $x$ and $y$ have the same \emph{pumping decomposition} if $x = uv^iw$ and $y = uv^jw$ for some $u$, $v$, $w$, $i$ and $j$.

\paragraph{Compression}
Compression refers to the practice of encoding information using fewer units of information (e.g.\ \emph{bits}) than the original representation. 
The function that provides such an encoding is called a \emph{compressor}.
We formally define a compressor as a so-called `code word length function' $C : \Sigma^{\ast} \to \mathbb{N}$ that maps strings to the length of their compressed encoding.
A normal compressor can be used to approximate the \emph{Kolmogorov complexity} \cite{ming1997introduction} of a string (i.e. the length of the shortest computer program  that produces the string as output), as the Kolmogorov complexity itself is not computable.
In \cite[Definition 3.1 and Lemma 3.4]{cilibrasi2005clustering}, a \emph{normal compressor} is defined as follows.

\begin{definition}
A compressor $C$ is \emph{normal} if it satisfies, up to an additive $O(log\ n)$ term, with $n$ the maximal length of an element involved in the (in)equality concerned, the following axioms:
\begin{enumerate}
  \item $C(uu)=C(u)$ and $C(\epsilon)=0$ (idempotency)
  \item $C(uv) = C(vu)$ (symmetry)
  \item $C(uv) \leq C(u) + C(v)$ (subadditivity)
\end{enumerate}
\end{definition}

The \emph{normalized compression distance} (NCD) is a quasi-universal measure for the similarity of two strings.
It is an approximation of their (uncomputable) \emph{information distance}, which measures the minimum information required to go from one string to the other or vice versa.
The NCD for two strings $x, y$ is defined in \cite{cilibrasi2005clustering} as follows.
\begin{equation}
NCD(x,y) = \frac{max\{C(xy)-C(x),C(yx)-C(y)\}}{max\{C(x),C(y)\}}
\end{equation}

\vspace*{0.1cm}
\section{Clustering by Compression}
\vspace*{0.1cm}

The NCD can be used to cluster strings based on their pumping decomposition.

\paragraph{Pumping lemma and clustering based on NCD}

Let us consider a setting where $\cal{L}$ and $\cal{L'}$ are two disjoint regular languages, i.e. $\cal{L} \cap \cal{L'} = \emptyset$. 
Let $x,y \in \cal{L}$ two strings such that $x = uv^iw$ and $y = uv^jw$ with $i,j \geq 0$ and $i \neq j$. 
Let now $z \in \cal{L'}$ be a string with $z = rs^kt$, $k \geq 0$ and $(r \neq u$ or $s \neq v$ or $t \neq w)$. 
We note here that $\cal{L}$ and $\cal{L'}$ are special cases of \textit{simple} regular languages in which any string belonging to the language can be fully described by the star-free regular expression of the language, together with the multiplicities of the repetitions. 

We now consider a proper normal compressor $C$, which would reduce any string in $\cal{L}$ or $\cal{L'}$ into the respective regular expressions of $\cal{L}$ and $\cal{L'}$, up to an additive $O(log\ n)$ term (the number of bits needed to encode the multiplicities).

Thus, we can say that, up to an additive $O(log\ n)$ term, $\forall x, y \in$ $\cal{L}$ $\ \ C(x) = C(y) = C(regexp($$\cal{L}$$))$

We also assume that the normal compressor we use satisfies, up to an additive $O(log\ n)$ term, the equality $C(xy) = C(x)$, since a proper normal compressor $C$ would compress equally the concatenation of two similar strings, i.e. belonging to the same simple language, and the strings separately. Since we know that $C(x) = C(y)$, and that $C(xy) = C(yx)$ by symmetry, we can then reformulate the NCD between $x$ and $y$ as 
\[NCD(x,y) = \frac{C(xy)-C(x)}{C(x)} = 0\]

As in our setting, $\cal{L}$ and $\cal{L'}$ are two disjoint regular languages not sharing the same regular expressions, we can thus assume that
\[
NCD(x, y) \leq NCD(x, z) \quad \textrm{and} \quad NCD(x, y) \leq NCD(y, z)
\]

By considering pairwise disjoint regular languages, we can say that a clustering based on the NCD will regroup sufficiently long strings in the format defined above and belonging to the same language. 

In practice, note that the additive $O(log\ n)$ term has an impact on the purity of the clustering. 
We will therefore carry out different clusterings of strings, possibly corresponding to clear distinct languages, depending on the granularity of the clustering, and of the number of clusters to find. 

Given a set of strings, the pairwise NCDs are used to compute a distance matrix. 
The next step is to establish clusters from the previously computed distance matrix. 
The output of hierarchical clustering is a dendrogram, a binary tree with one leaf for every string.
A dendrogram is composed of internal nodes, linking leaves or subtrees. 
The level of the nodes is set up according to a pre-defined linkage matrix computed as a function of the distance matrix.

\begin{figure}[!t]
	\centering
    \hspace*{-0.2cm}
    \vspace*{-1.0cm}
	\includegraphics[height=0.39\textheight]{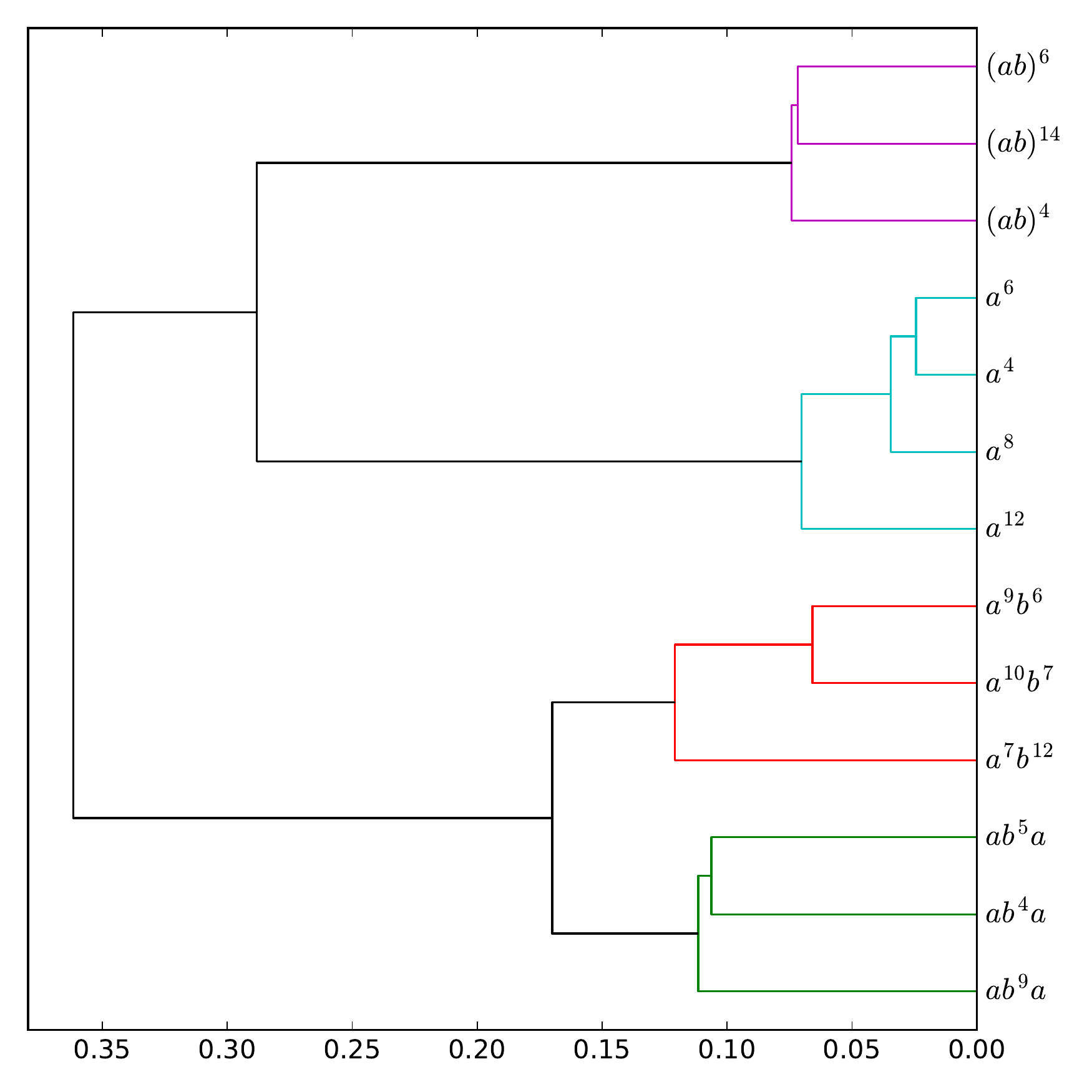}
    \vspace*{0.1cm}
	\caption[Clustering of a sample of the printers dataset.]{Clustering of a sample of the printers dataset.}
    \vspace*{-0.4cm}
	\label{fig:clusteringexample}
\end{figure}

To build the dendrogram, hierarchical agglomerative clustering algorithms find and merge the pair of clusters with the lowest linkage, and create a parent node at the level of the linkage. 
We gather the dendrogram, and thus clusters are computed by varying the level of linkage as a threshold. 
We assume that the number of languages to get, thus the right threshold to set, is known beforehand.

As shown in the example in Figure \ref{fig:clusteringexample}, clusters are created and highlighted with different colors, assuming a threshold of $0.15$ for the separation. 
In this case, we gather 4 clusters, from which it is possible to infer distinct languages. 
It is straightforward here to see that, for a given cluster $\cal{C}$$_x$, we can infer a language $\cal{L}$$_x$ such that $\forall i \geq 0,\ uv^iw \in \cal{L}$$_x$. 
In order to do so, it is possible for any cluster $\cal{C}$$_x$ to infer the corresponding language $\cal{L}$$_x$ (for example, by using state merging \cite{cdlh}), using all strings clustered in $\cal{C}$$_x$ as positive examples, and all strings in $\bigcup_y \mathcal{C}_y : y \neq x$ as negative examples.
We then gather 4 languages $\cal{L}$$_{1\ldots 4}$ respectively standing for $a^+$ (blue), $a^+b^+$ (red), $ab^+a$ (green) and $(ab)^+$ (purple).

\paragraph{Granularity of Clustering}
As shown above, choosing the right threshold enables getting the desired number of clusters. As a consequence, we assume that the right number of clusters is provided, in order to enhance the identification of different languages. In case the wrong number of clusters would have been provided, or an inadequate threshold found, the granularity of the clustering, that is to say, how fine the strings are clustered together, would have changed. More precisely, a greater threshold would tend to generalize the optimal solution (a solution with a threshold of $0.2$ would give 3 languages, namely $a^+$, $(ab)^+$ and $a^+b^+a^{[0..1]}$) whereas a smaller threshold would provide finer languages (up to singleton sets).

\section{Tandem repeats}

Another approach is to first compute the pumping decompositions of the strings (and regroup the strings accordingly) by computing \emph{tandem repeats}.

 \begin{definition}% A string $w$ is a \textit{tandem array} if it can written as $w = a^k$ for some $k \geq 2$. In the case $k = 1$, $w$ is called a \textit{primitive}.
A string $v$ is a \emph{tandem repeat} in $x$ if $x = uv^iw$ for some $i \geq 2$, $u,w \in \Sigma^*$ and $v \neq \epsilon$.

\end{definition}

There exist algorithms that can detect tandem repeats in a string \cite{gusfield2004linear,crochemore1981optimal,apostolico1983optimal,main1984n}.
We relate the pumping lemma for regular languages to the identification of tandem repeats, in a context of simple regular languages.

\paragraph{Pumping lemma and identification of languages}

According to the pumping lemma, if $\cal{L}$ is a regular language, then there exists an integer $p \geq 1$ depending only on $\cal{L}$ such that every string $x \in \cal{L}$ of length at least $p$ can be written as $x = uvw$ such that $|v| \geq 1$, $|uv| \leq p$ and $\forall i \geq 0,\ uv^iw \in \cal{L}$.
If, in addition $\mathcal{L}$ is a simple language, then there is only a finite number of such decompositions.
Therefore, if we assume an upper bound on the number of states of the DFA for $\cal{L}$, two sufficiently long strings belong to the same language if they contain a similar prefix, tandem repeat, and suffix.
 
Algorithms for finding tandem repeats in a string can thus be used to cluster strings.
Let us consider the string $x = uvvw$. 
Tandem repeat finder algorithms will naturally find $v$ as being a tandem repeat. 
As a consequence, we can use the tandem repeat found to describe the string as $x = uv^2w$. 
We can then generalize this to infer the simple language $\{uv^iw : i \geq 2\}$.

\begin{table*}[!t]
\begin{center}
\begin{tabular}{lcccccc}
\toprule
$abc$ 					&  \multicolumn{3}{c}{$abc$}  & \multirow{3}{*}{$(abc)^*$} & \multirow{10}{*}{$\Sigma^*$} \\
\cline{1-4}
$abcabc$ 				& \multicolumn{2}{c}{\multirow{2}{*}{$(abc)^{\geq 2}$}} & \multirow{2}{*}{$(abc)^+$} \\
\cline{1-1}
$abcabcabc$ 			&   \\
\cline{1-5}
$ababab$ 				& \multicolumn{2}{c}{\multirow{2}{*}{$(ab)^{\geq 2}$}} & \multirow{2}{*}{$(ab)^+$} & \multirow{7}{*}{$(a^*b^*)^*$} \\
\cline{1-1}
$abababab$ 				&   \\
\cline{1-4}
$aabaabaab$  			& \multirow{2}{*}{$(aaab)^{\geq 2}$} & \multirow{2}{*}{$(a^{\geq 2}b)^{\geq 2}$} & \multirow{2}{*}{$(a^+b)^+$} &  \\
\cline{1-1}
$aabaabaabaabaab$ 		\\
\cline{1-4}
$aabbbaabbbaabbb$ 		& $(aabbb)^{\geq 2}$ & \multirow{2}{*}{$(a^{\geq 2}b^{\geq 2})^{\geq 2}$} & \multirow{2}{*}{$(a^+b^+)^+$} &  &  \\
\cline{1-2}
$aaabbaaabbaaabb$ 		& $(aaabb)^{\geq 2}$  \\
\cline{1-3}
$aaabbaaabbaabbbaabbb$  & $(aaabb)^{\geq 2}(aabbb)^{\geq 2}$ & $(a^{\geq 2}b^{\geq 2})^{\geq 2}(a^{\geq 2}b^{\geq 2})^{\geq 2}$  \\

\bottomrule
\end{tabular}
\vspace*{0.1cm}
\caption{Clustering of languages, from fine grained solutions (left) to coarse (right).}
\label{table:clusteringlanguages}
\end{center}
\vspace*{-1cm}
\end{table*}

Now, let us consider a string $x = abcbcbcbcd$. 
String $x$ contains two distinct tandem repeats, namely $bc$ and $bcbc$. 
By taking the longest tandem repeat that is not a tandem repeat itself, we can say $x = a(bc)^4d$. 
By generalizing, we can infer the simple language $\{a(bc)^id : i \geq 2\}$.

Tandem repeats can also be computed recursively. 
Let us consider the string $x = aaabcbcaaabcbcaaabcbc$. 
We can first detect the tandem repeat $aaabcbc$, and state that $x = (aaabcbc)^3$. 
If we again process the result, we can detect tandem repeats $a$ and $bc$, and describe $x$ as $x = (a^3(bc)^2)^4$. 
By generalizing, we can infer the simple language $\{(a^i(bc)^j)^k : i,j,k \geq 2\}$.

\paragraph{Ambiguities}
Let $x$ be the string $abaabaab$. 
Then, $x$ contains several possible branching tandem repeats, namely $t_1 = baa$, $t_2 = aab$ or $t_3 = aba$. 
Thus, we can describe the language for $x$ in different ways, namely $\mathcal{L}_1 = a(baa)^{\geq 2}b$, $\mathcal{L}_2 = ab(aab)^{\geq 2}$ or $\mathcal{L}_3 = (aba)^{\geq 2}ab$. 
Nevertheless, all these representations are equal, i.e. $\forall i \geq 0$, $a(baa)^ib = ab(aab)^i = (aba)^iab$. 
Therefore, we claim that any of these representations is valid in order to infer the language.
It does not matter which one choose, as long as we check for equivalence. 

\paragraph{Hierarchical clustering of languages}
We have defined several steps to generalize pumping decompositions of strings based on tandem repeats. 
As shown above, we can describe strings as $x = uv^iw$ for all $i \geq 2$, and cluster them into $uv^{\geq 2}w$. 
We then propose to lower the multiplicity of $v$ to $\geq 1$, to get $uv^+w$. 
The next step of generalization is to get $uv^*w$ by lowering the multiplicity of $v$ to $\geq 0$. 
We can then imagine $\Sigma^*$ to be an ultimate generalization.
Indeed, $\{uvvw\} \subset uv^{\geq 2}w \subset uv^+w \subset uv^*w \subset \Sigma^*$. 
In this generalization process, some strings might share some generalizations from given steps. 
In Table \ref{table:clusteringlanguages} we propose such a hierarchical clustering of regular languages, from singleton sets to $\Sigma^*$.

\section{Industrial Application}

\begin{table}[!b]
\vspace*{-0.5cm}
\begin{center}
\begin{tabular}{lcc}
\toprule
printjob & pattern & \hspace{0.5cm} type of printjob \\
\midrule
$\ aaaaa$ 						& \multirow{3}{*}{$a^+$}	& \multirow{3}{*}{homogeneous job}	\\
\cline{1-1}
\rule{0pt}{2.5ex} $aaaaaaaaaa$ & 						 	&									\\
\cline{1-1}
\rule{0pt}{2.5ex} $aaaaa\ldots aaa$ &					&									\\
\cline{1-3}
\rule{0pt}{2.5ex} $abababab$	& \multirow{2}{*}{$(ab)^+$}	& \multirow{4}{*}{heterogeneous job}			\\
\cline{1-1}
\rule{0pt}{2.5ex} $abababababababab$	&					& 									\\
\cline{1-2}
\rule{0pt}{2.5ex} $abcabcabc$	& \multirow{2}{*}{$(abc)^+$}& 									\\
\cline{1-1}
\rule{0pt}{2.5ex} $abcabcabcabcabcabc$ & 					&									\\
\cline{1-3}
\rule{0pt}{2.5ex} $abcbcbcbca$ & $a(bc)^+a$ 				&\multirow{2}{*}{booklet}\\
\cline{1-2}
\rule{0pt}{2.5ex} $abcdebcdebcdea$	& $a^+(bcde)^+a^+$ 	& 									\\
\bottomrule
\end{tabular}
\vspace*{0.1cm}
\caption{Sample of identified print-job patterns.}
\label{table:patterns-example}
\end{center}
\vspace*{-0.5cm}
\end{table}
Our case study has been inspired by an industrial problem related to the domain of large-scale printers. 
Recent work \cite{waqas} focused on the impact of design parameters of an industrial printer on its productivity. 
It appeared in the aforementioned study that the productivity depends on the print-job being rendered. 
To that end, the prior identification of the different print-job patterns is crucial in enabling printer engineers to find out optimal parameters related to paper flow, according to each significant type of job being printed. 
It is also important to identify significant usages of the printers, in order to set up components as a function of these usages. 
Indeed, the productivity of large-scale printers is sensitive to changes of media types, since some component settings (such as printheads height or paper speed for instance) need to be readjusted each time a different media type is used. 
Thus, frequent changes of the type of sheets within a print job may lead to a decrease of productivity due to different component settings, if not correctly supported.  

We possess a dataset containing more than 23.000 distinct strings, each representing a print job. 
Each print job is composed of several pages, that can be of different media types (papers of different thickness, sizes, etc.). 
Thus, each string is composed of letters standing for the media-type of the printed page. 
We could have for instance $aaaaaabbbbbbbbbb$ that would represent a print job consisting of 6 pages of type $a$, and 10 pages of type $b$.

A characteristic of our dataset, and of our strings to cluster, is that the occurrence of new symbols in the strings follows alphabetical order. 
For instance, if we consider $\Sigma = \{a,b,c\}$, our strings will always start with an $a$, followed eventually by either an $a$ or a $b$, etc.
In Table \ref{table:patterns-example}, we present a sample of the significant print job patterns identified using the method involving tandem repeats.

\section{Conclusion and Further Work}

In this paper, we have described different heuristics for learning pairwise disjoint simple languages from positive examples. 
The results obtained within the scope of our industrial case study indicate that these methods are applicable in practice. 
Nevertheless, we realize that the success of our methods is related to the simplicity of the languages we learn. 
Therefore, in future work we aim to formally define a class of languages for which our methods work.

\section*{Acknowledgments}

We would like to thank Lou Somers and Patrick Vestjens\footnote{\{lou.somers,patrick.vestjens\}@oce.com} for providing industrial datasets as well as required expertise related to the case study.
This research is supported by the Dutch Technology Foundation (STW) under the Robust CPS program (project 12693) and by the Netherlands Organisation for Scientific Research (NWO) project on Learning Extended State Machine for Malware Analysis (LEMMA, project 628.001.009).

\bibliographystyle{IEEEtran}
\bibliography{lib}

\begin{thebibliography}{10}
\providecommand{\url}[1]{#1}
\csname url@samestyle\endcsname
\providecommand{\newblock}{\relax}
\providecommand{\bibinfo}[2]{#2}
\providecommand{\BIBentrySTDinterwordspacing}{\spaceskip=0pt\relax}
\providecommand{\BIBentryALTinterwordstretchfactor}{4}
\providecommand{\BIBentryALTinterwordspacing}{\spaceskip=\fontdimen2\font plus
\BIBentryALTinterwordstretchfactor\fontdimen3\font minus
  \fontdimen4\font\relax}
\providecommand{\BIBforeignlanguage}[2]{{%
\expandafter\ifx\csname l@#1\endcsname\relax
\typeout{** WARNING: IEEEtran.bst: No hyphenation pattern has been}%
\typeout{** loaded for the language `#1'. Using the pattern for}%
\typeout{** the default language instead.}%
\else
\language=\csname l@#1\endcsname
\fi
#2}}
\providecommand{\BIBdecl}{\relax}
\BIBdecl

\bibitem{Gold1967}
\BIBentryALTinterwordspacing
E.~M. Gold, ``Language identification in the limit,'' \emph{Information and
  Control}, vol.~10, no.~5, pp. 447 -- 474, 1967. [Online]. Available:
  \url{http://dx.doi.org/10.1016/S0019-9958(67)91165-5}
\BIBentrySTDinterwordspacing

\bibitem{rabin1959finite}
M.~O. Rabin and D.~Scott, ``Finite automata and their decision problems,''
  \emph{IBM journal of research and development}, vol.~3, no.~2, pp. 114--125,
  1959.

\bibitem{gusfield2004linear}
D.~Gusfield and J.~Stoye, ``Linear time algorithms for finding and representing
  all the tandem repeats in a string,'' \emph{Journal of Computer and System
  Sciences}, vol.~69, no.~4, pp. 525--546, 2004.

\bibitem{ming1997introduction}
L.~Ming and P.~Vit{\'a}nyi, \emph{An introduction to Kolmogorov complexity and
  its applications}.\hskip 1em plus 0.5em minus 0.4em\relax Springer
  Heidelberg, 1997.

\bibitem{cilibrasi2005clustering}
R.~Cilibrasi and P.~M. Vit{\'a}nyi, ``Clustering by compression,'' \emph{IEEE
  Transactions on Information theory}, vol.~51, no.~4, pp. 1523--1545, 2005.

\bibitem{cdlh}
C.~de~la Higuera, \emph{Grammatical inference: learning automata and
  grammars}.\hskip 1em plus 0.5em minus 0.4em\relax Cambridge University Press,
  2010.

\bibitem{crochemore1981optimal}
M.~Crochemore, ``An optimal algorithm for computing the repetitions in a
  word,'' \emph{Information Processing Letters}, vol.~12, no.~5, pp. 244--250,
  1981.

\bibitem{apostolico1983optimal}
A.~Apostolico and F.~P. Preparata, ``Optimal off-line detection of repetitions
  in a string,'' \emph{Theoretical Computer Science}, vol.~22, no.~3, pp.
  297--315, 1983.

\bibitem{main1984n}
M.~G. Main and R.~J. Lorentz, ``An o (n log n) algorithm for finding all
  repetitions in a string,'' \emph{Journal of Algorithms}, vol.~5, no.~3, pp.
  422--432, 1984.

\bibitem{waqas}
U.~Waqas, M.~Geilen, S.~Stuijk, J.~v.~Pinxten, T.~Basten, L.~Somers, and
  H.~Corporaal, ``A fast estimator of performance with respect to the design
  parameters of self re-entrant flowshops,'' in \emph{Euromicro Conference on
  Digital System Design}, Aug 2016, pp. 215--221.

\end{thebibliography}

\end{document}